\documentclass[conference]{IEEEtran}
\IEEEoverridecommandlockouts
\usepackage{cite}
\usepackage{amsmath,amssymb,amsfonts,amsthm,mathtools}
\usepackage[alphabetic,backrefs,lite,nobysame]{amsrefs}
\usepackage{algorithmic}
\usepackage{graphicx}
\usepackage{textcomp}
\usepackage{xcolor}
\usepackage{subcaption}

\def\BibTeX{{\rm B\kern-.05em{\sc i\kern-.025em b}\kern-.08em
		T\kern-.1667em\lower.7ex\hbox{E}\kern-.125emX}}

\newtheorem{theorem}{Theorem}[section]

\theoremstyle{definition}

\newtheorem{definition}[theorem]{Definition}

\numberwithin{figure}{section}

\begin{document}
	
	\title{Geometric Fusion via Joint Delay Embeddings\\
		\thanks{Both authors were partially supported by the Air Force Office of Scientific Research under grant AFOSR FA9550-18-1-0266. We are grateful to Erik Blasch for technical guidance, and to Kenneth Ball, John Harer, Jay Hineman, Tessa Johnson, Gary Koplik, and Lihan Yao for helpful discussions.}
	}
	
	\author{\IEEEauthorblockN{1\textsuperscript{st} Elchanan Solomon}
		\IEEEauthorblockA{\textit{Department of Mathematics,} \\
			\textit{Duke University}\\
			Durham, USA \\
			yitzchak.solomon@duke.edu}
		\and
		\IEEEauthorblockN{2\textsuperscript{nd} Paul Bendich}
		\IEEEauthorblockA{\textit{Department of Mathematics,} \\
			\textit{Duke University}\\
			\textit{Geometric Data Analytics}\\
			Durham, USA \\
			paul.bendich@duke.edu}
	}
	
	\maketitle
	
	\begin{abstract}
		We introduce geometric and topological methods to develop a new framework for fusing multi-sensor time series. This framework consists of two steps: (1) a joint delay embedding, which reconstructs a high-dimensional state space in which our sensors correspond to observation functions, and (2) a simple orthogonalization scheme, which accounts for tangencies between such observation functions, and produces a more diversified geometry on the embedding space. We conclude with some synthetic and real-world experiments demonstrating that our framework outperforms traditional metric fusion methods.

	\end{abstract}
	
	
	\section{Introduction}
	
	Data fusion \cite{blasch2012} is the task of synthesizing measurements made by different sensors and sensing modalities, and is broadly divided into two paradigms: downstream fusion, in which the synthesis happens after each sensor has arrived at a classification or regression decision, and upstream fusion \cite{UpstreamDataFusion}, where the synthesis happens on the level either of the raw measurements or features extracted from the raw measurements. We are interested in the task of upstream fusion, in particular for time series.
	
	To motivate our particular approach to time series fusion, consider the following real-world example. A hurricane passes over Durham Country, NC, and the local weather station receives temperature, precipitation, wind strength, humidity, and barometric time series. Intuitively, we understand that a model of the full dynamics of the hurricane system is the ideal fusion of these related measurements. By analogy, given any collection of time series, our goal is to reconstruct a dynamical system in a higher-dimensional state space for which our time series come from observation functions, measuring statistics about the state of the system as it evolves.
	
	In addition to its easy interpretability, this dynamical systems approach can call upon a rich, mathematical literature, in particular the study of delay embeddings. Delay embeddings provide a method for reconstructing higher-dimensional dynamics from the time series of a single observation function. We define \emph{joint delay embeddings} as a natural extension of this technique to multiple time series, and propose it as a solution to this upstream fusion task.
	
	One challenge in building our state space is that its geometry is dependent on the correlations between our observation functions. Returning to our hurricane example, elementary physics tells us that the pressure and temperature of a gas increase proportionally to one another. If we do not take this into account, we will record an increase in temperature as being distinct from an increase in pressure, inflating the dimensionality of the state space, and introducing a bias into the geometry, wherein it appears the system is growing at twice the rate it actually is. To address this lack of independence between observation functions, we define a one-parameter family of operators, which we call \emph{Gram-Schmidt tensors}, which take as input a set of vectors in Euclidean space, partially orthogonalize them, and sum their norms.
	
	Finally, we demonstrate that a combination of Joint Delay Embeddings with Gram-Schmidt tensors produces superior results to other geometric fusion algorithms for some synthetic and real-world data sets.
	
	
	\subsection{Outline}
	
	The rest of this paper proceeds as follows. Delay embeddings and their relation to the \emph{topology} of state space, as well as our novel concept of \emph{joint} delay embedding, are discussed in Section \ref{sec:DETT}. Then Section \ref{sec:GDE} introduces the problem of reconstructing the \emph{geometry} of state space and proposes our solution. The concept of metric fusion, with two particular manifestations from the literature, appears in Section \ref{sec:MF}.
	Finally, we show the benefits of our approach via experiments with synthetic data in Section \ref{sec:SE}, and real data in Section \ref{sec:MDS}.
	
	\section{Delay Embeddings and Takens' Theorem}
	\label{sec:DETT}
	We now review the concept of a delay embedding and outline Takens' foundational embedding theorem. Given a metric space $\mathcal{O}$, consider a continuous time series $x(t):\mathbb{R} \to \mathcal{O}$. For the sake of concreteness, the reader can take $\mathcal{O}$ to be $\mathbb{R}$, but the generality of our construction allows for higher dimensional Euclidean spaces, function spaces, and even metric spaces of shapes. Concrete examples include acoustic time series, moving vehicles, multi-sensor EEG scans, videos, graphs with time-dependent edge weights, etc. Given a delay parameter $\tau$ and a dimension parameter $d$, one can construct a time series $X_{\tau,d}(t)$ in $\mathcal{O}^d$ called the \emph{delay embedding} or \emph{sliding window embedding}, defined as follows:
	\[X_{\tau,d}(t) = (x(t), x(t + \tau), x(t + 2\tau), \cdots, x(t + (d-1)\tau))\]
	The name \emph{delay embedding} corresponds to the fact that $X_{\tau, d}(t)$ records the value of the time series at time $t$, as well as at $d$ delayed, future times. Interest in delay embeddings is motivated by the following, theoretical result of Takens:
	
	\begin{theorem}[\cite{takens1981detecting}]
		Let $M$ be an $n$-dimensional manifold, and $f:M \to M$ a generic smooth diffeomorphism.  Let $\alpha: M \to \mathbb{R}$ be a smooth and suitably generic observation function, and consider the $k$-dimensional embedding $\phi: M \to \mathbb{R}^k$ given by $\phi_{k}(x) = (\alpha(x),\alpha(f(x)), \cdots, \alpha(f^{k-1}(x)))$. Then, if $k \geq 2n+1$, $\phi_{k}$ is an embedding of $M$ into $\mathbb{R}^k$.
	\end{theorem}
	
	To interpret this in the setting of data analysis, we may think of $M$ as the state space of a complex, hidden system, and $f: M \to M$ as a discrete evolution map, describing how the state evolves from one moment in time to the next. The observation function $\alpha$ maps a state $p \in M$ to a single, real-valued statistic $\alpha(p)$, such as the temperature of a solution, the bearing of a ship, or the intensity of a sound wave (living in the state spaces of a chemical reaction, naval trajectory, and musical performance, respectively). If we fix an initial point $p_0 \in M$, we obtain a discrete time series $x(n):\mathbb{N} \to \mathbb{R}$:
	\[x(n) = \alpha(\overbracket{f \circ f \circ \cdots \circ f}^{\mbox{n times}}(p_0)).\]
	
	Because the observation function $\alpha$ only outputs a single, real value, it will often fail to be injective, i.e. there will be distinct states $p,q \in M$ with $\alpha(p) = \alpha(q)$. In order to distinguish such states, we can consider the value of $\alpha$ at their successor states, $f(p)$ and $f(q)$. By replacing the scalar value $\alpha(p)$ with the vector $(\alpha(p),\alpha(f(p)))$, we remove some of the failures of injectivity, at the cost of obtaining a time-series in $\mathbb{R}^2$. However, it may still be the case that there exist distinct states $p,q \in M$ with $(\alpha(p),\alpha(f(p))) = (\alpha(q),\alpha(f(q)))$, so we can go further and consider $f(f(p)), f(f(f(p)))$, and so on. Takens' theorem assures us that we need to consider at most $2\operatorname{dim}(M)+1$ future time steps to completely resolve any lack of injectivity, and so obtain a topologically accurate reconstruction of $M$ in $\mathbb{R}^{2\operatorname{dim}(M)+1}$. The utility of such a reconstruction is that it provides an interpretable and structured summary of the data that reveals features that are otherwise either invisible or the output of unwieldy, black-box trained models. Turning this perspective around, if we start with a time series $x(t):\mathbb{R} \to \mathbb{R}$ which we believe comes from an observation function on a higher-dimensional state space, the delay embedding will build a representation of an orbit in that state space. This construction naturally generalizes to time series in any metric space, as outlined above. We point the reader to a rich literature (e.g, \cite{perea2015sliding}, \cite{Swipe}, and \cite{tralie2016high}) about using sliding window embeddings to study the topology of data.

	\subsection{Joint Delay Embeddings}
	
	Consider next a set of time series $x_{1}(t), \cdots, x_{m}(t)$, each valued in a distinct metric space $\mathcal{O}_1, \cdots, \mathcal{O}_m$, as arises in the setting of multisensor fusion. For example, a recording of a person using microphones and cameras of various resolutions will be valued in distinct discretizations of audio or pixel space, respectively. We can construct a delay embedding $X_{i}(t)$ for each time series separately, but this only sees the dynamics from the perspective of a single observation function. Instead, we propose a \emph{joint delay embedding}:
	\begin{definition}[Joint Delay Embedding]
		Let $x_{1}(t), \cdots, x_{m}(t)$ be a finite collection of time series. Given a delay parameter $\tau$ and dimension parameter $d$, the joint delay embedding (JDE) of our time series is a time series of $m \times d$ matrices\footnote{This is a slight abuse of terminology, as the entries of these matrices need not be numbers, but are elements in a variety of metric spaces.} $\mathbf{X}(t)$ with elements $\mathbf{X}(t)_{ij} = x_{i}(t + (j-1)\tau)$. 
	\end{definition}
	In principle, the joint delay embedding $\mathbf{X}(t)$ requires a smaller dimension parameter $d$ than any of the individual delay embeddings $X_{1}(t), \cdots, X_{n}(t)$, due to the larger number of observation functions used. Moreover, and more importantly, the resulting embedding fuses the information of these time series in an interpretable and non-trivial way.
	
	Before moving on, we point out two features of our definition of joint delay embeddings:
	1) The definition of a joint delay embedding can be modified to allow for time-warping phenomena between time series. All that is required is to replace the fixed delay parameter $\tau$ with a variable parameter keyed to local time-warping effects. For the sake of improving readability, we ignore this subtlety, with the understanding that the following analysis is substantially identical in the more general setting.
	2) A time series in $\mathbb{R}^k$ can be split into $k$ time series in $\mathbb{R}$. For the moment, this makes no difference to the resulting joint delay embedding. However, it does have a subtle effect on the geometric constructions of the next section. Put concisely, we do not consider tangencies between one-dimensional time series once they are packaged together into a single $\mathbb{R}^k$ time series. Thus, even if all the component time series are identical, we consider them to contain orthogonal information.

	\section{Geometry of Delay Embeddings}
	\label{sec:GDE}
	Takens' theorem asserts that the delay time-series $\phi$ is an embedding, and hence preserves the \emph{topological} structure of the manifold $M$. However, it tells us nothing about how the geometry of the embedding relates to the geometry of the state manifold. An accurate representation of the true, underlying geometry is important for distinguishing noise from signal, as well as for the effective application of most machine learning models, which are sensitive to the scale and magnitude of the input vectors. Returning to our recording example, we do not want to produce a state manifold that is dominated by the speech of whichever person is the focus of the majority of the audio or video sensors. In general, determining the geometry of $M$ from one, or many, observation functions is not a well-posed problem, especially when we do not have any constraints on the distortion of those functions. However, there are some heuristics we can adopt that help narrow down the solution space. The hypothesis we propose here is that it is very unlikely, in the absence of highly symmetric data, for two independent observation functions to produce delay embeddings with similar geometries. Thus, if we observe such similarities, we can rescale our state manifold accordingly, avoiding the distortion associated with recording the same feature multiple times. Let us now make this precise.
	
	
	
	Given a finite collection of time series $x_{1}(t), \cdots, x_{m}(t)$, consider the joint delay embedding with parameters $d$ and $\tau$. For a pair of times $t_1$ and $t_2$, we can form vectors $w_{i}$ in $\mathbb{R}^d$ as follows:
	\[w_{i} = (d_{\mathcal{O}_i}(x_{i}(t_1),x_{i}(t_2)), d_{\mathcal{O}_i}(x_{i}(t_1 + \tau),x_{i}(t_2 + \tau)),\]
	\[ \cdots
	d_{\mathcal{O}_i}(x_{i}(t_1 + (d-1)
	\tau),x_{i}(t_2 + (d-1)\tau))  )\]  
	
	That is, $w_i$ is the difference between the delay embedding vectors $X_{i,\tau,d}(t_1)$ and $X_{i,\tau,d}(t_2)$. If we were to assume all our observation functions were independent, we would define the distance between the joint delay embedding vectors $\mathbf{X}(t_1)$ and $\mathbf{X}(t_2)$ to be $\sqrt{\|w_1\|^2 + \cdots + \|w_{m}\|^2}$. However, if our observation functions are not independent, this would give undue weight to repeated observations, and minimize the impact of unique ones. To modify this distance in a way that takes into account the angles between the vectors $w_1, \cdots, w_m$, we introduce the following tensor:
	
	

	\begin{definition}[Gram-Schmidt Tensor]
		Let $V = \{w_1, \cdots, w_m\}$ be a collection of vectors, and $\lambda \in [0,1]$ an orthogonality parameter. We define $N_{\lambda}(V)$ to be the real value produced by the following algorithm.
		\begin{enumerate}
			\item Initialize the algorithm with all vectors unmarked.
			\item Let $w^{*}$ be the unmarked vector with largest $\ell^2$ norm. Mark $w^*$.
			\item For all vectors $w \neq w^{*}$, replace $w$ with $w - \lambda  \frac{\langle w,w^{*} \rangle}{\langle w^{*},w^{*} \rangle} w^{*}$.
			\item If there are remaining unmarked vectors, return to step (2). Otherwise, proceed to the next step.
			\item Return $\sqrt{\|w_1\|^2 + \cdots + \|w_m\|^2}$.
		\end{enumerate}
	\end{definition}
	
	When $\lambda = 1$, the above algorithm reduces the collection $V$ to a set of orthogonal vectors, as in the Gram-Schmidt algorithm. However, when $\lambda < 1$, we enforce orthogonality less strictly, and the resulting value $N_{\lambda}(V)$ is larger than $N_{1}(V)$. We can also consider the role of $\tau$ and $d$ in $N_{\lambda}(V)$: as the delay vector grows, we have more data with which to determine orthogonality, affecting the result of our above algorithm, and as the size of the delay parameter $\tau$ grows, the comparison of our observation functions becomes more global.
	
	
	\section{SNF and JDL}
	\label{sec:MF}
	Before considering the results of Joint Delay Embeddings on a synthetic data set, we review two well-known techniques in metric fusion, against which the performance of JDE will be evaluated. The first technique is Similarity Network Fusion (SNF, \cite{wang2014similarity}). As the name suggests, SNF is designed for fusing similarities, rather than distances, so necessitates a pre-processing step of turning distance matrices into similarity matrices. The second technique is Joint Distance Learning (JDL), a modification of Joint Manifold Learning in the context of discrete metric spaces.
	
	\subsection{The SNF Algorithm}
	The following pipeline and equations follow the treatment of SNF as outlined by Tralie, Bendich, and Harer in \cite{tralie2019multi}. The input to the SNF algorithm is a collection of distance matrices. We first transform our distance matrices into similarity matrices as follows. If $D = (D_{ij})$ is one such matrix, we set:
	\[W_{ij} = \exp \left( - \frac{D_{ij}^2}{\Sigma_{ij}}  \right),\]
	
	where $\Sigma_{ij}$ is a parameter that measures the average distances from $x_i = x(t_i)$ and $x_j = x(t_j)$ to nearby points. If this is small, the exponential decays faster, and hence only the closest points have large similarity. If it is large, the exponential decays more slowly, and further away points still have significant similarity. The formula for $\Sigma_{ij}$ is as follows, and depends on two constants: $\kappa \in [0,1]$ and $\beta \in \mathbb{R}_{>0}$:
	
	\[\Sigma_{ij}^{\kappa} = \frac{\beta}{3}\left(\frac{1}{\kappa N}\left(\sum_{k \in N^{\kappa}(i)} D_{ik} \right) + \frac{1}{\kappa N}\left(\sum_{k \in N^{\kappa}(j)} D_{jk} \right) + D_{ij}     \right).\] 
	The set $N^{\kappa}(i)$ is obtained by ordering the points $\{x_1, \cdots, x_N\}$ in order of decreasing distance from $x_i$ and taking the first $\kappa$ percent of these points. Thus $N^{\kappa}(i)$ consists of the first $\kappa N$ nearest neighbors of $x_i$. The next step is to build two normalizations of $W_{ij}$. The first normalization is:
	\[P_{ij} = \begin{cases}
	\frac{W_{ij}}{2\sum_{k \neq i}W_{ik}} & j \neq i\\
	1/2 & j=i
	\end{cases}.\]
	This has the effect of normalizing the rows of $W_{ij}$, so that they sum to $1$. The second normalization is similar, but is only supported on pairs of nearest neighbors. Let $N^{i}$ denote $N^{\kappa}(i) \setminus \{x_i\}$. We define:
	\[S_{ij} = \begin{cases}
	\frac{W_{ij}}{2\sum_{k \in N_{i}}W_{ik}} & j \in N_{i}\\
	1/2 & j=i\\
	0 & \mbox{otherwise}.
	\end{cases}\]
	
	Now, suppose we have distance matrices $D^{1}, \cdots, D^{m}$, and thus similarity matrices $W^{1}, \cdots, W^{m}$ and normalizations $P^{1}, \cdots, P^{m}$ and $S^{1}, \cdots, S^{m}$. The SNF algorithm consists of iteratively updating the $P$ matrices in order of ascending index. That is, for $l$ ranging from $1$ to $m$, we define:
	
	\[P^{l} = S^{l} \times \frac{\sum_{k \neq l}P^k}{m-1} \times (S^l)^T.\]
	
	Updating all the $P$ matrices constitutes one step of the SNF algorithm. We can repeat this process a number of times, cycling through the $P$ matrices. Finally, we average the outputs to produce:
	\[\bar{P} = \frac{1}{m}\sum_{l=1}^{m}P^{l}. \]
	
	The matrix $\bar{P}$ is the output of the SNF algorithm, and can be thought of as a fusion of the original $P$ matrices.
	
	\subsection{Joint Distance Learning}
	Joint Manifold Learning (JML), introduced by Davenport et. al. in \cite{davenport2010high} and developed further by Shen et al. in \cite{ShenJML2018}, is a technique for fusing a family of $K$-dimensional manifolds $\{M_1, \cdots, M_J\}$ homeomorphic to a fixed, model manifold $M$ via homeomorphisms $\psi_{i}:M_{i} \to M$. One considers the product manifold $\mathbf{M} = M_1 \times \cdots \times M_J$, and then restricts to the joint manifold: $M^{*} \subset \mathbf{M}$: $M^{*} = \{(\psi_{1}(p), \cdots, \psi_{J}(p)) \mid p \in M  \}$. The output of JML is the intrinsic, geodesic distance on the joint manifold $M^*$. Although JML is useful in a variety of contexts, it is poorly suited to fusing metrics on time series. This is because the geodesic metric on a curve is blind to its embedding in space, and so a non self-intersecting curve wrapping around a torus or sphere inherits the same geometry as a straight line. Thus, the limitation of JML is not the manifold assumption, which can be relaxed, but the use of the geodesic metric. To that end, we propose the following, extrinsic adaption of JML.
	
	\begin{definition}[Joint Distance Learning]
		Let $d_1, \cdots, d_n$ be a finite collection of metrics on a common set $X$. The joint distance learning (JDL) fusion of these metrics is defined as:
		\[\hat{d}^{2}(x_1,x_2) = d_{1}^{2}(x_1,x_2) + \cdots + d_{n}^{2}(x_1,x_2)\]
	\end{definition}  
	
	The common approach taken in JML and JDL is to view the set of geometries observed as orthogonal slices of a single geometry embedded in high-dimensional space. We can therefore view JDL as a special case of JDE, with $d$ and $\lambda$ both being equal to $0$.
	
	\section{Synthetic Experiments}
	\label{sec:SE}
\begin{figure}
		\begin{center}
			\includegraphics[scale=0.35]{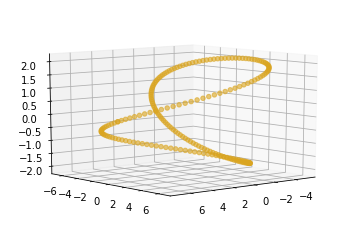} \includegraphics[scale=0.35]{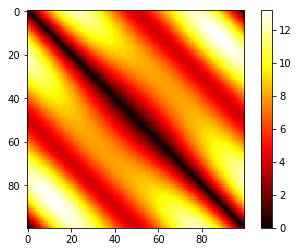}
			\includegraphics[scale=0.35]{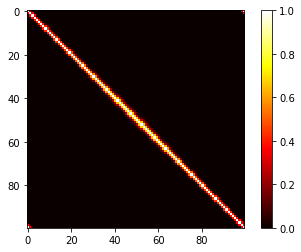}
			
			\caption{The $(1,2)$ torus curve and its associated Euclidean distance matrix and similarity matrix $(\beta = 0.5, \kappa = 0.1$).}
			\label{fig:toruscurve}
		\end{center}
	\end{figure}

	In this section, we consider three experiments in multi-sensor fusion, and compare the results of JDE, SNF and JDL. For all experiments, we work with the same synthetic data set, for which we have access to a ground-truth distance matrix. Our set $X$ of data points consists of $N=100$ points evenly spaced along the parameterized space curve:
	\[x(t) = (R + r \cos(ax + x_0 ))\cos(bx + y_0)\]
	\[y(t) = (R + r \cos(ax + x_0 ))\sin(bx + y_0)\]
	\[z(t) = r \sin(ax + x_0), \]
	
	where $(x_0,y_0) = (0,0)$, $(R,r) = (5,2)$, and $(a,b) = (1,2)$. This is a curve on a torus with outer radius $R=5$ and inner radius $r=2$. The curve starts at the point $(0,0)$ and loops once around the meridian and twice around the longitude of the torus; see Figure \ref{fig:toruscurve} for a visualization of this curve. We can compute a distance matrix whose entries are the pairwise Euclidean distance between points on this curve, as well as the associated similarity matrix, see Figure \ref{fig:toruscurve}. 
	
	Our experiments are the following: (1) We uniformly pick three random vectors $v_1, v_2, v_3$ on the sphere, and define time series via the coordinate functions $\alpha_{i}(x) = \langle x, v_i \rangle$, (2) We uniformly pick three random points $p_1, p_2, p_3$ in the box $[-2.5,2.5]^3$, and define time series via the coordinate functions $\alpha_{i}(x) = \| x - p_i\|$, (3) We take four sensors, two given by random projections, and two given by distances to random basepoints.  Our goal, for every experiment, is to produce a fused distance matrix that most closely resembles the ground truth.
	
	\subsection{Experiment (1)}
\begin{figure}
		\begin{center}
			\includegraphics[scale=0.3]{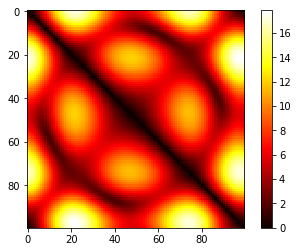} \includegraphics[scale=0.3]{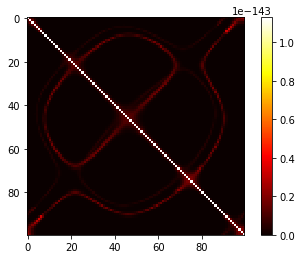}
			\includegraphics[scale=0.3]{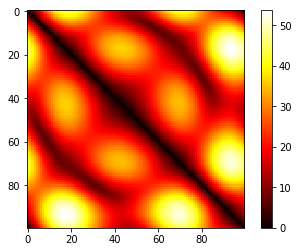}
			\includegraphics[scale=0.3]{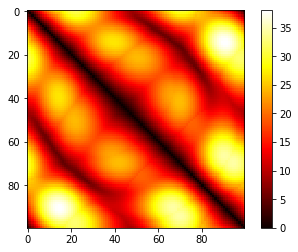}
			\includegraphics[scale=0.3]{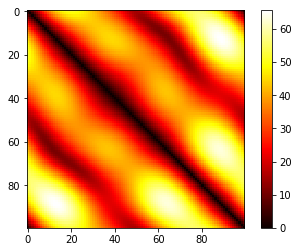}
			\includegraphics[scale=0.3]{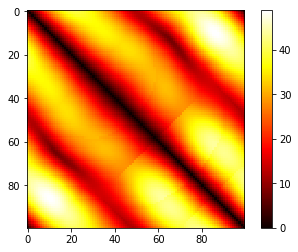}

			\caption{Top Left: JDL. Top Right: SNF $(\beta = 0.5, \kappa = 0.1$). Center Left: JDE $(d=10,\lambda=0)$. Center Right: JDE $(d=10,\lambda=1)$. Bottom Left: JDE $(d=20,\lambda=0)$. Bottom Right: JDE $(d=20,\lambda=1)$.}
			\label{fig:Experiment1}
		\end{center}
	\end{figure}

	We compare the results of JDL, SNF, and four variants of JDE, with $d$ equal to $10$ or $20$, and $\lambda$ equal to $0$ and $1$. See Figure \ref{fig:Experiment1}. We observe that JDL and SNF do a poor job of reconstructing the correct distance or similarity matrix. For JDE with $d=10$, the reconstruction of the distance matrix is much improved, especially when the orthogonality parameter $\lambda$ is set to $1$. Finally, when $d=20$, JDE does well at both values of $\lambda$, with $\lambda=1$ only providing a slightly better result.
	
	\subsection{Experiment (2)}
\begin{figure}
		\begin{center}
			\includegraphics[scale=0.25]{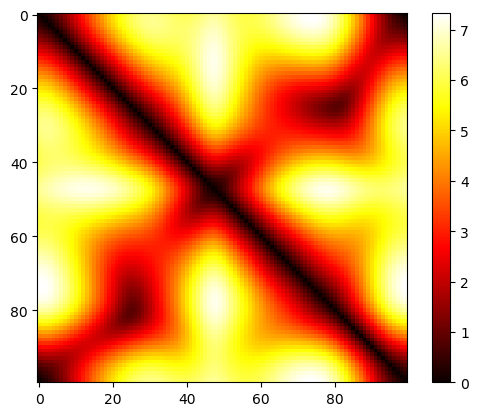} \includegraphics[scale=0.25]{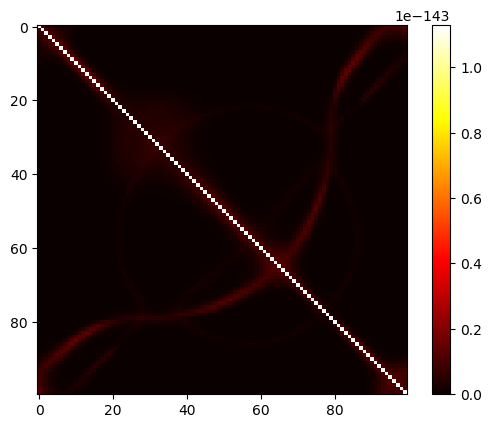}
			\includegraphics[scale=0.25]{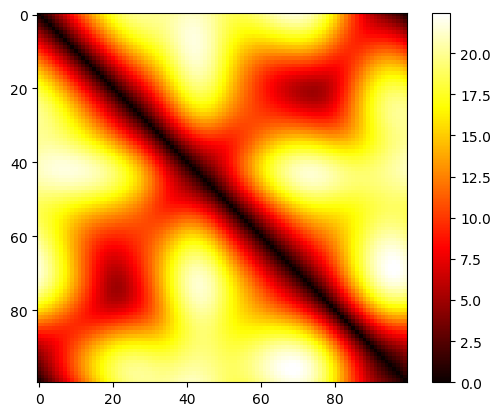}
			\includegraphics[scale=0.25]{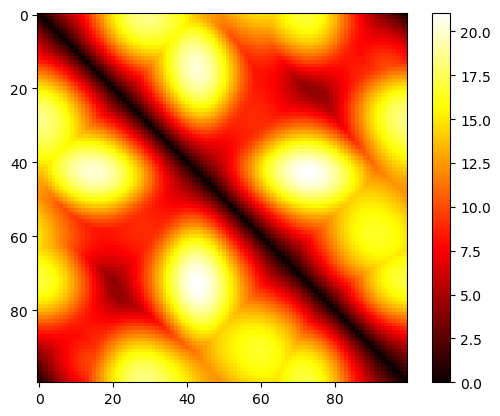}
			\includegraphics[scale=0.25]{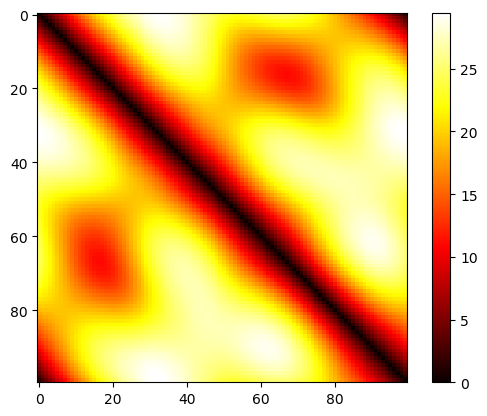}
			\includegraphics[scale=0.25]{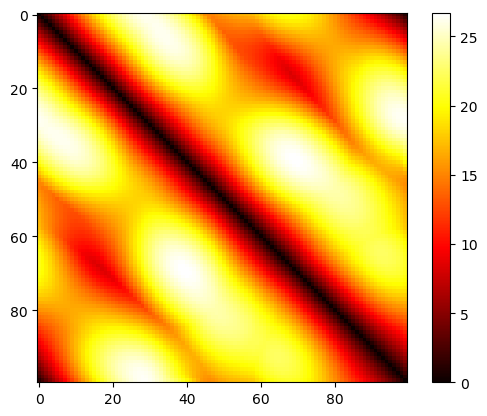}

			\caption{Top Left: JDL. Top Right: SNF $(\beta = 0.5, \kappa = 0.1$). Center Left: JDE $(d=10,\lambda=0)$. Center Right: JDE $(d=10,\lambda=1)$. Bottom Left: JDE $(d=20,\lambda=0)$. Bottom Right: JDE $(d=20,\lambda=1)$.}
			\label{fig:Experiment2}
		\end{center}
	\end{figure}

	As in the prior experiment, the performance of JDL and SNF is quite poor. However, the results of JDE now demonstrate the significant impact of the orthogonality parameter $\lambda$, as the $\lambda=1$ fusion far outperforms the $\lambda=0$ fusion in both $d=10$ and $d=20$. See Figure \ref{fig:Experiment2}.
	\subsection{Experiment (3)}
\begin{figure}
		\begin{center}
			\includegraphics[scale=0.3]{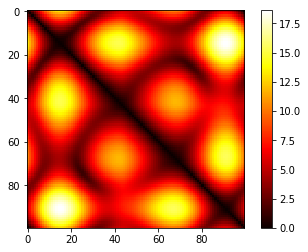} \includegraphics[scale=0.3]{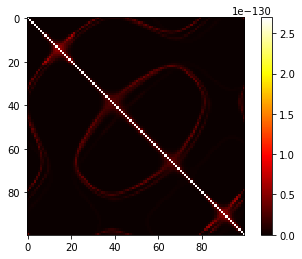}
			\includegraphics[scale=0.3]{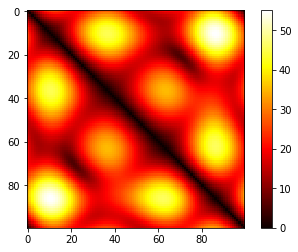}
			\includegraphics[scale=0.3]{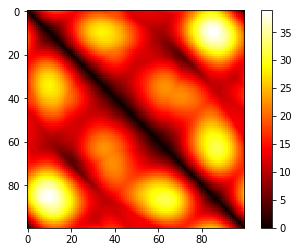}
			\includegraphics[scale=0.3]{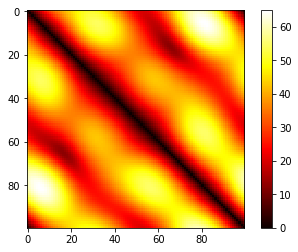}
			\includegraphics[scale=0.3]{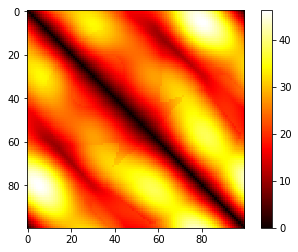}

			\caption{Top Left: JDL. Top Right: SNF $(\beta = 0.5, \kappa = 0.1$). Center Left: JDE $(d=10,\lambda=0)$. Center Right: JDE $(d=10,\lambda=1)$. Bottom Left: JDE $(d=10,\lambda=0)$. Bottom Right: JDE $(d=10,\lambda=1)$.}
			\label{fig:Experiment3}
		\end{center}
	\end{figure}

	Finally, in the third experiment, we again see that JDL and SNF produce distorted fusions, and that JDE with $\lambda=1$ gives the best results. See Figure \ref{fig:Experiment3}.
	
	\section{MotionSense Data Set}
	\label{sec:MDS}
	
	The MotionSense Data Set, coming from the work of Malekzadeh et. al. in \cite{malekzadeh2019mobile}, consists of smartphone accelerometer and gyroscope measurements recorded for 24 individuals performing 6 activities in 15 trials in the same environment and conditions: going downstairs, going upstairs, walking, jogging, sitting, and standing. There are 12 modalities measured by the smartphones: attitude.roll, attitude.pitch, attitude.yaw, gravity.x, gravity.y, gravity.z, rotationRate.x, rotationRate.y, rotationRate.z, userAcceleration.x, userAcceleration.y, userAcceleration.z. Our goal is to apply time series fusion as a means of doing unsupervised learning, by identifying meaningful geometric structure in the state space of human motion. As JDE and JDL give distance matrices as output, we compare them in two ways:
	\begin{enumerate}
		\item Using Multi-Dimensional Scaling (MDS), a method for embedding a discrete metric space in low-dimensional Euclidean space that approximately preserves its geometry.
		\item Using persistence diagrams \cite{Edelsbrunner2000}, a method from topological data analysis that extracts salient, structural features, such as clusters and cycles, from discrete shapes. Note that these features are extracted directly from the high-dimensional embeddings, rather from than the low-dimensional approximations provided by MDS.
	\end{enumerate}
	
\begin{figure}
		\begin{center}
			\includegraphics[scale=0.3]{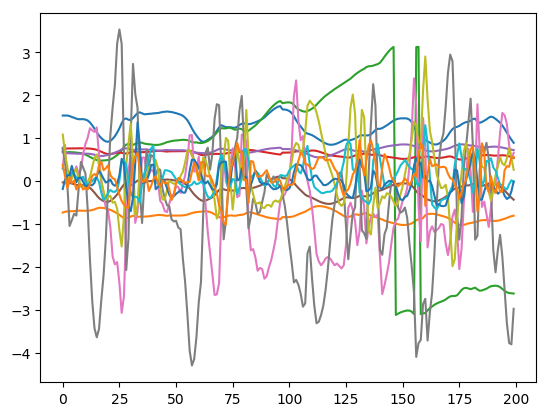} \includegraphics[scale=0.3]{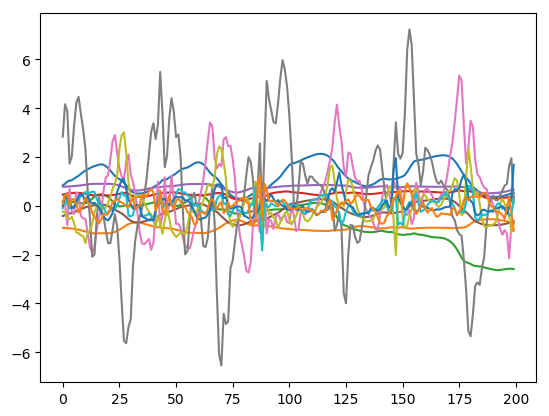}
			\includegraphics[scale=0.3]{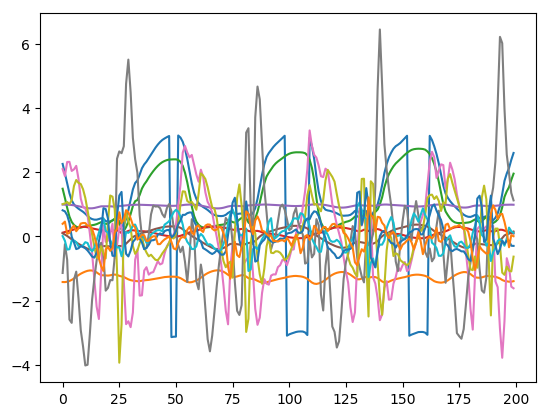}
			\includegraphics[scale=0.3]{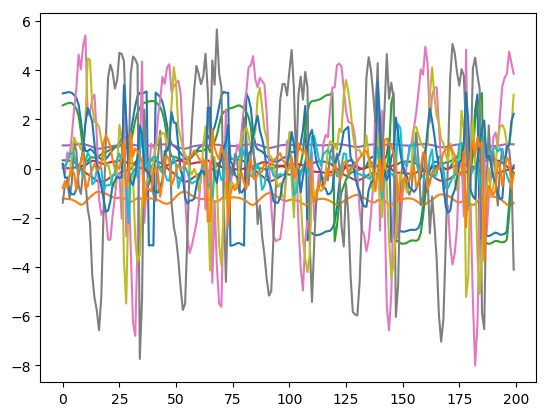}
			
			\caption{Smartphone measurement time series. Top Left: Downstairs. Top Right: Upstairs. Bottom Left: Walking. Bottom Right: Jogging.}		
			\label{fig:timeseries}
		\end{center}
	\end{figure}

	As we consider the results for different fusion methods, we ask ourselves the same question: ``is there any meaningful structure or pattern observable using this method that cannot be detected using the the other methods?" Due to the relative complexity of our analysis, we focus on a single participant (the first one), considering the first $200$ time steps for four tasks: going downstairs (``downstairs 1"), going upstairs (``upstairs 12"), walking (``walking 8"), and jogging (``jogging 9"). See Figure \ref{fig:timeseries} for plots of the raw time series, in which the complexity of our fusion task is readily apparent.

	\subsection{Downstairs}
\begin{figure}
		\begin{center}
			\includegraphics[scale=0.3]{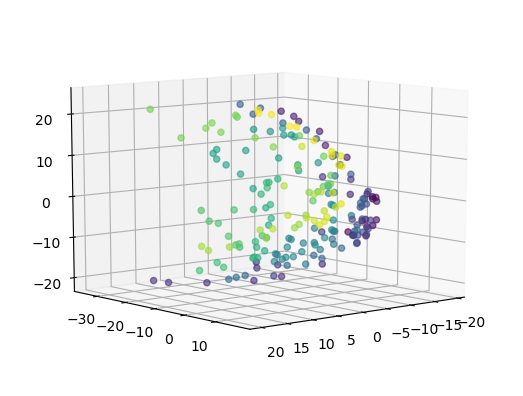} \includegraphics[scale=0.25]{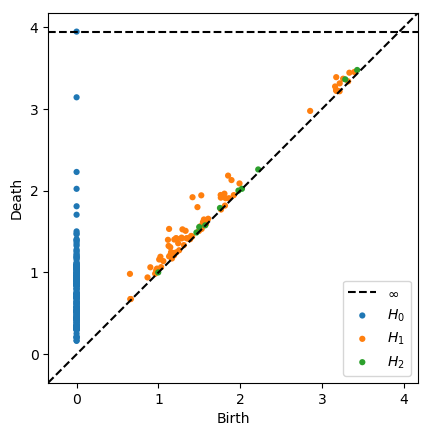}
			\includegraphics[scale=0.3]{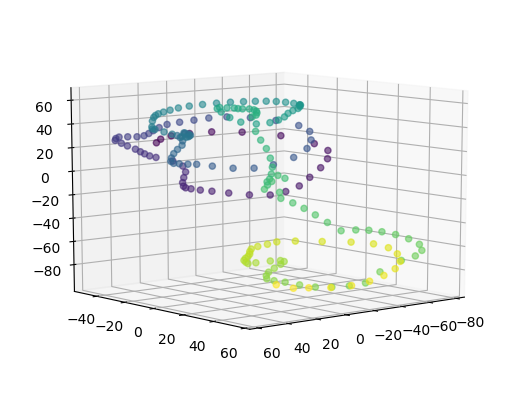}
			\includegraphics[scale=0.25]{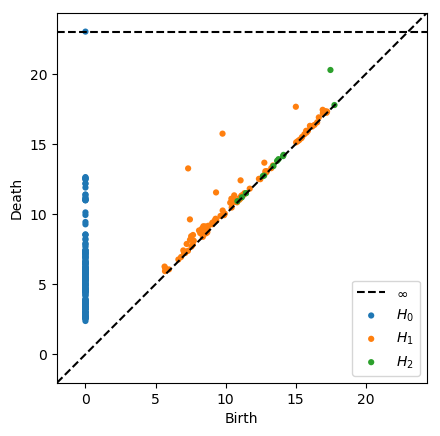}
			\includegraphics[scale=0.3]{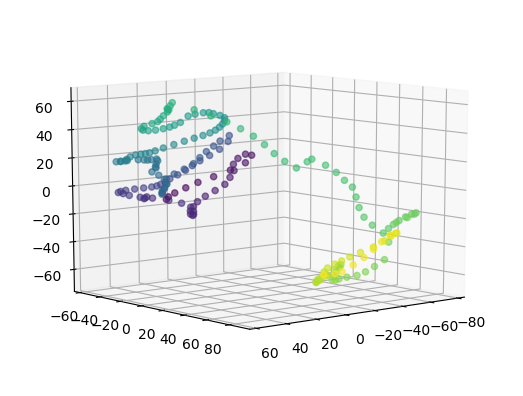}
			\includegraphics[scale=0.25]{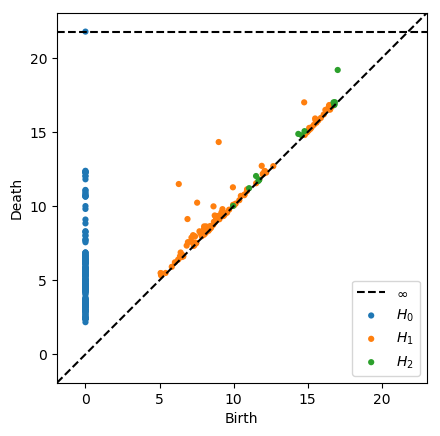}

			\caption{Downstairs. Top Left: MDS Embedding of JDL fusion. Top Right: Persistence Diagram of JDL fusion. Middle Left: MDS Embedding of JDE $(d=20,\lambda = 0)$. Middle Right: Persistence Diagram of JDE $(d=20,\lambda = 0)$. Bottom Left: MDS Embedding of JDE $(d=20,\lambda = 1)$. Bottom Right: Persistence Diagram of JDE $(d=20,\lambda = 1)$.}
			\label{fig:dws1}
		\end{center}
	\end{figure}

See Figure \ref{fig:dws1}. We see that the MDS plot of the JDL fusion is noisy, with little apparent structure. Similarly, the persistence diagram of the JDL fusion does not present interesting higher-dimensional features. By contrast, the JDE fusion MDS shows a time series that loops around before wrapping around a sphere, and the corresponding persistence diagrams contain 1-dimensional (orange) persistence points, corresponding to the loops, and a 2-dimensional (green) persistence point, corresponding to the sphere. The advantage of the $\lambda = 1$ fusion over the $\lambda = 0$ fusion, as seen in the MDS plots, is that the geometry of the sphere is more accurate. The authors speculate that the emergence of this sphere is related to the nature of the smartphone measurements, as pitch, yaw, and rotation are tied to spherical geometry.
	
	\subsection{Upstairs}
\begin{figure}
		\begin{center}
			\includegraphics[scale=0.3]{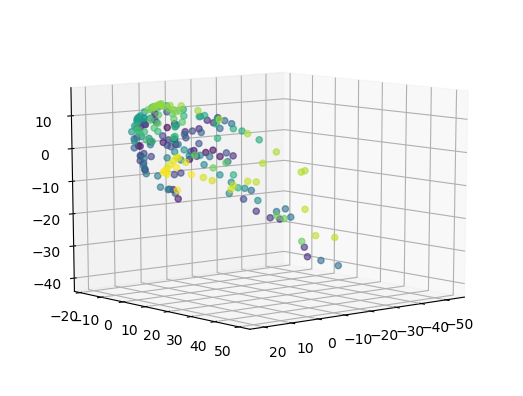} \includegraphics[scale=0.3]{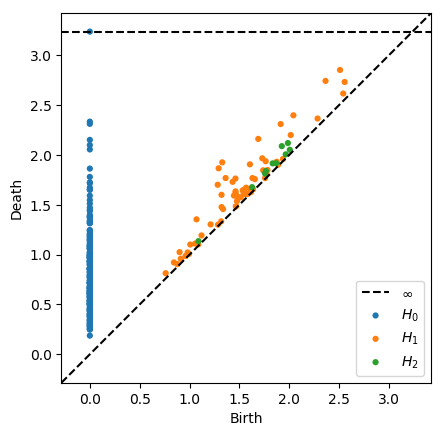}
			\includegraphics[scale=0.3]{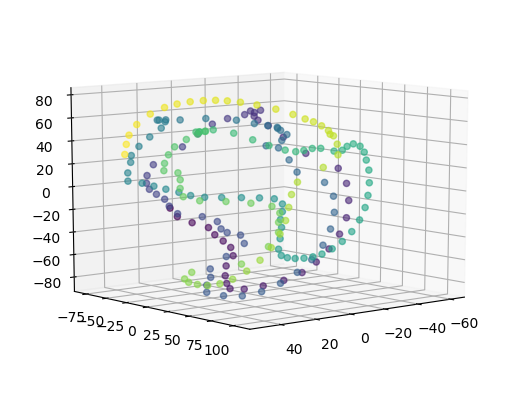}
			\includegraphics[scale=0.3]{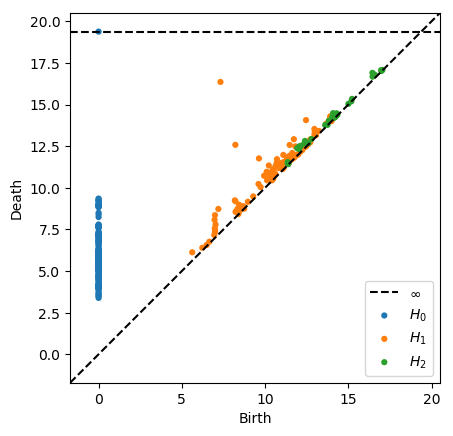}
			\includegraphics[scale=0.3]{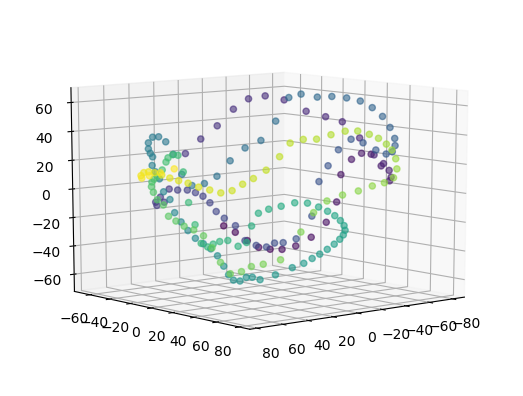}
			\includegraphics[scale=0.3]{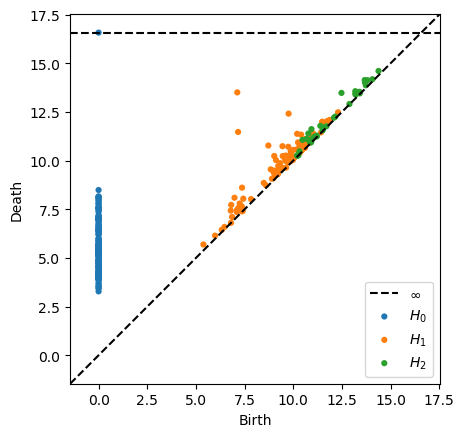}

			\caption{Upstairs. Top Left: MDS Embedding of JDL fusion. Top Right: Persistence Diagram of JDL fusion. Middle Left: MDS Embedding of JDE $(d=20,\lambda = 0)$. Middle Right: Persistence Diagram of JDE $(d=20,\lambda = 0)$. Bottom Left: MDS Embedding of JDE $(d=20,\lambda = 1)$. Bottom Right: Persistence Diagram of JDE $(d=20,\lambda = 1)$.}
			\label{fig:ups3}
		\end{center}
	\end{figure}

See Figure \ref{fig:ups3}. As before, the MDS plot of the JDL fusion is noisy, with little apparent structure. Similarly, the persistence diagram of the JDL fusion does not present interesting higher-dimensional features. The JDE fusion with $\lambda = 0$ exhibits an interesting quasi-periodic structure, with a single one-dimensional persistent point far from the diagonal, and one more a middling distance away. The JDE fusion with $\lambda = 1$ shows one more one-dimensional persistence point far from the diagonal, as well as a two-dimensional persistence point near the diagonal that may be noise, or may correspond to a torus, a hypothesis matched by the associated MDS plot. 
	
	\subsection{Walking}
\begin{figure}
		\begin{center}
			\includegraphics[scale=0.3]{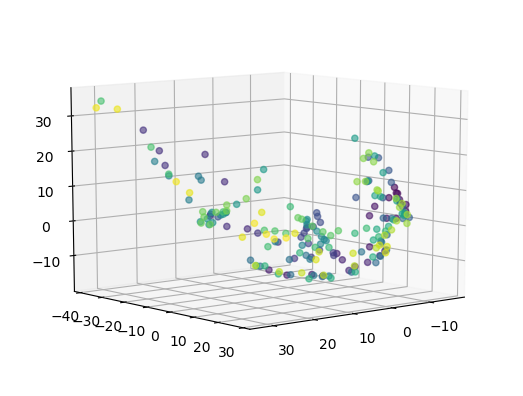} \includegraphics[scale=0.3]{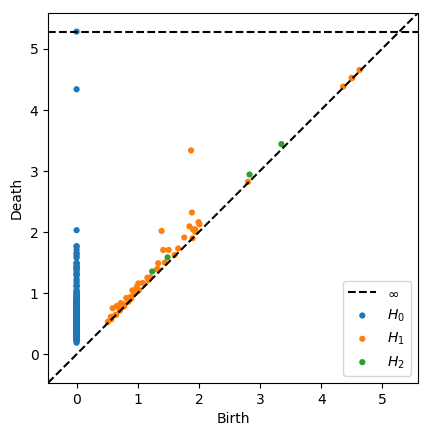}
			\includegraphics[scale=0.3]{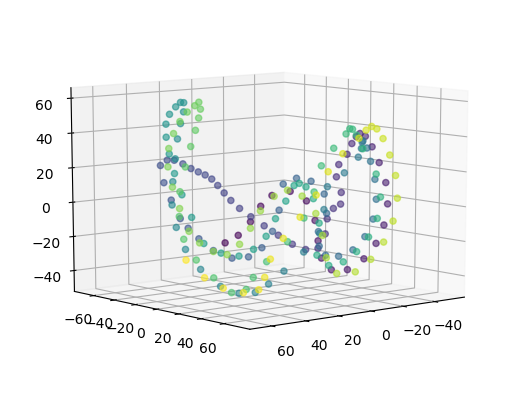}
			\includegraphics[scale=0.3]{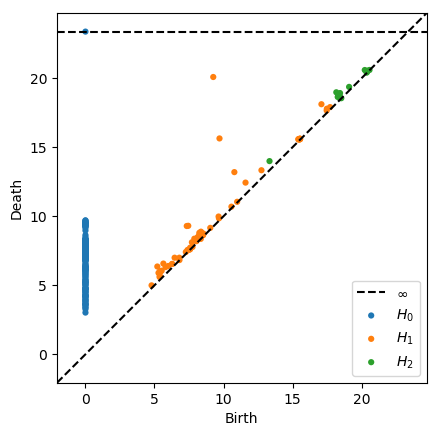}
			\includegraphics[scale=0.3]{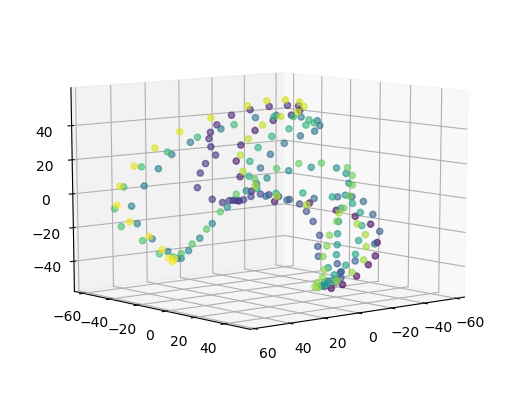}
			\includegraphics[scale=0.3]{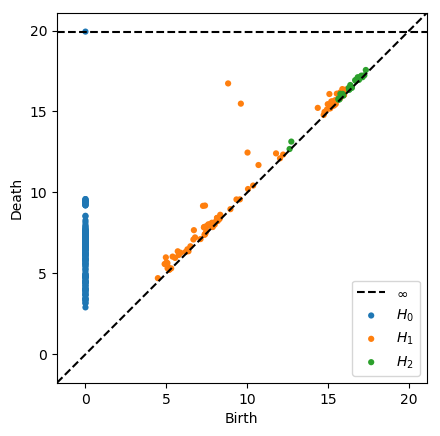}

			\caption{Walking. Top Left: MDS Embedding of JDL fusion. Top Right: Persistence Diagram of JDL fusion. Middle Left: MDS Embedding of JDE $(d=20,\lambda = 0)$. Middle Right: Persistence Diagram of JDE $(d=20,\lambda = 0)$. Bottom Left: MDS Embedding of JDE $(d=20,\lambda = 1)$. Bottom Right: Persistence Diagram of JDE $(d=20,\lambda = 1)$.}
			\label{fig:wlk8}
		\end{center}
	\end{figure}

See Figure \ref{fig:wlk8}. The MDS plot of the JDL fusion is noisy, and the persistence diagram captures a single one-dimensional cycle far from the diagonal. The JDE fusions produce more structured MDS plots and reveal more one-dimensional persistence points, especially for $\lambda = 1$.
	
	\subsection{Jogging}
\begin{figure}
		\begin{center}
			\includegraphics[scale=0.3]{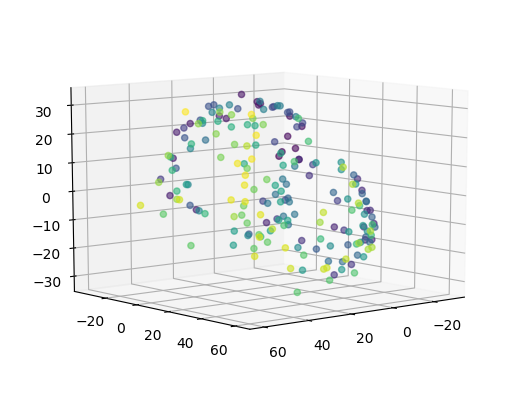} \includegraphics[scale=0.3]{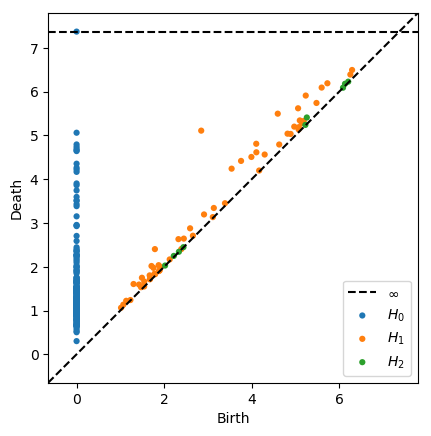}
			\includegraphics[scale=0.3]{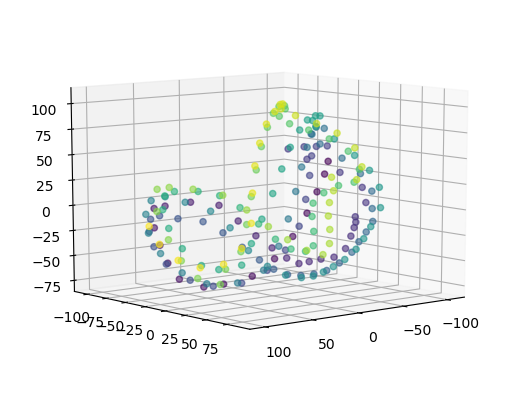}
			\includegraphics[scale=0.3]{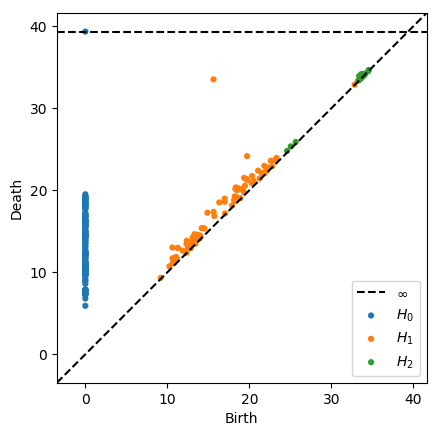}
			\includegraphics[scale=0.3]{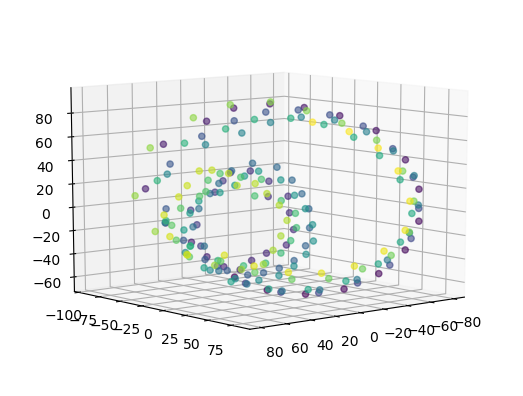}
			\includegraphics[scale=0.3]{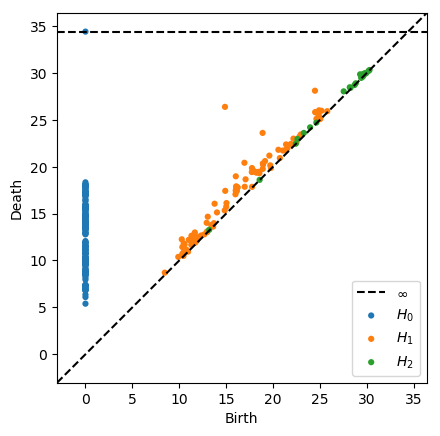}

			\caption{Jogging. Top Left: MDS Embedding of JDL fusion. Top Right: Persistence Diagram of JDL fusion. Middle Left: MDS Embedding of JDE $(d=20,\lambda = 0)$. Middle Right: Persistence Diagram of JDE $(d=20,\lambda = 0)$. Bottom Left: MDS Embedding of JDE $(d=20,\lambda = 1)$. Bottom Right: Persistence Diagram of JDE $(d=20,\lambda = 1)$.}
			\label{fig:jog9}
		\end{center}
	\end{figure}

See Figure \ref{fig:jog9}. As in the walking trial, MDS plot of the JDL fusion is noisy, and the persistence diagram captures a single one-dimensional cycle far from the diagonal. The JDE fusions produce more structured MDS plots and reveal more one-dimensional persistence points, especially for $\lambda = 1$.
	
	Altogether, we see that JDE fusion produces more structured and interpretable results as compared to JDL, and that this is further enhanced via our orthogonalization scheme. A more thorough geometric and topological analysis of the MotionSense data set is beyond the scope of this article, however, and will be the subject of future work.
	
	\section{Conclusion} 
	
	Our central thesis in this article was that the ideal fusion of a family of time series is a reconstruction of their unified state space. We argued that a joint delay embedding successfully accomplishes this fusion, but may not produce the most accurate or informative geometry. To that end, we introduced Gram-Schmidt tensors as a way of correcting for local tangencies between sensors. Our synthetic and real-world experiments demonstrated that a combination of joint delay embeddings with Gram-Schmidt tensors can outperform other metric fusion methods in the literature.
	
	With regards to future research, there are theoretical and applied directions of interest. In the former area, we would like to provide precise guaranties on the accuracy of our geometric reconstructions as depends on the distortion of the individual observation functions. In the applied arena, the investigation of the MotionSense data set seems promising, as does the analysis of EEG data, which also fits neatly into our framework.

	\bibliographystyle{plain}
	\bibliography{fusionbib}
	
	%
	
\end{document}